\documentclass[runningheads]{llncs}
\usepackage[T1]{fontenc}
\usepackage{graphicx}
\usepackage{hyperref, xcolor}
\hypersetup{colorlinks=true, urlcolor=blue, linkcolor=black, citecolor=black}

\usepackage{tabulary}
\usepackage{float}
\usepackage{listings}
\usepackage{caption}
\usepackage{subcaption}
\usepackage{multirow}
\usepackage{tcolorbox}
\usepackage{booktabs}
\usepackage{comment}
\usepackage{natbib}
\tcbset{
    exactbox/.style={
        colframe=black!50,        
        colback=gray!5,           
        sharp corners=south,      
        boxrule=0.6pt,            
        leftrule=1mm,             
        arc=1mm,                  
        width=\textwidth,         
        left=2pt, right=1pt, top=2pt, bottom=2pt, 
        fontupper=\fontsize{5.2}{6.2}\selectfont    
        \fontfamily{phv}\selectfont
        
    } 
}

\begin{document}
%

\title{Baba is LLM: Reasoning in a Game with Dynamic Rules}%
%
\author{
Fien van Wetten\inst{1} \and
Aske Plaat\inst{1} \and
Max van Duijn\inst{1}}
\authorrunning{F. van Wetten et al.}
%
\institute{LIACS, Leiden University, The Netherlands}

\maketitle              

\begin{abstract}
Large language models (LLMs) are known to perform well on language tasks, but struggle with reasoning tasks.
%
This paper explores the ability of LLMs to play the 2D puzzle game \textit{Baba is You}, in which players manipulate rules by rearranging text blocks that define object properties. 
%
Given that this rule-manipulation relies on language abilities \textit{and} reasoning, it is a compelling challenge for LLMs.
%
Six LLMs are evaluated using different prompt types, including (1) simple, (2) rule-extended and (3) action-extended prompts. In addition, two models (Mistral, OLMo) are finetuned using textual and structural data from the game.
%
Results show that while larger models (particularly GPT-4o) perform better in reasoning and puzzle solving, smaller unadapted models struggle to recognize game mechanics or apply rule changes. Finetuning improves the ability to analyze the game levels, but does not significantly improve solution formulation. We conclude that even for state-of-the-art and finetuned LLMs, reasoning about dynamic rule changes is difficult (specifically,  understanding  the use-mention distinction).
%
The results provide insights into the applicability of LLMs to complex problem-solving tasks and highlight the suitability of games with dynamically changing rules for testing  reasoning  and reflection by LLMs.

\end{abstract}
\section{Introduction}


Artificial Intelligence (AI) has a long history in using games as benchmarks for reasoning, decision-making, and problem-solving capabilities~\citep{campbell2002deep,silver2016mastering,silver2018general,brown_superhuman_2018,openai_dota_2019,schrittwieser2020mastering}. This paper investigates the use of large language models (LLMs) in the 2D puzzle game \textit{Baba is You}~\citep{teikari2019baba}. In this game players must alter rules by manipulating text blocks. Solving puzzles in this environment requires understanding how rule changes affect the game state and to apply that understanding dynamically, implying a form of reasoning in which the model should be able to reflect on the effects of its own actions.

LLMs, based on the transformer architecture~\citep{vaswani_attention_2023} have demonstrated strong performance in natural language processing tasks including text generation, machine translation, conversational agents and code generation~\citep{naveed_comprehensive_2024}. Techniques such as finetuning~\citep{xu_parameter-efficient_2023}, reinforcement learning with human feedback (RLHF)~\citep{chaudhari_rlhf_2024}, and prompt-based learning~\citep{kamath_prompt-based_2024} have been developed to improve the performance of the transformer architecture. Beyond natural language processing, LLMs are also emerging as agents in games~\citep{meta2022human,topsakal2024evaluating,marincioni2024effect,muller2023chatter}.


\textit{Baba is You} is compelling because of its dynamic rule system relying on two-level language-based mechanics, which can be understood in terms of the classical \textit{mention} versus \textit{use} distinction \citep{wilson2017bridge,saka1998quotation}. Board games in general are based on pushing around pieces associated with a fixed meaning (mention). However, in \textit{Baba is You} certain pieces can form a new game rule when they are aligned (use). Unlike games with fixed rules, \textit{Baba is You} allows players to  rewrite the logic of the game by manipulating the pieces. While LLM's strong language and general pattern-learning abilities suggest potential~\citep{mirchandani_large_2023}, an initial study by~\citet{cloos2024baba} showed indeed that state-of-the-art LLMs struggle with the reasoning aspects of \textit{Baba is You}, failing to generalize rule manipulation.
 %

This paper  evaluates how well LLMs  solve \textit{Baba is You} puzzles. We use two approaches: prompt-based learning and finetuning. We test six LLMs (GPT-4o, 
Gemini-Flash 1.5, 
OLMo 2 13B and 7B, 
Mistral 7B and Mixtral 8x7B) 
across three prompt types. We additionally finetune Mistral 7B and OLMo 7B using game data. 
Our contributions are as follows:\begin{itemize}
    \item Comparing different prompts in six LLMs, we find that prompt-based learning achieves weak results when dynamic rule changes are necessary---even for LLMs with enhanced reasoning capacities;
    \item Using a dataset for finetuning, we find that finetuning on two open LLMs is able to improve performance somewhat;
    \item Reasoning about dynamic rules changes, remains a challenging problem for current Reasoning LLMs; the deceptively simple puzzle game of {\em Baba is You} offers a challenging testbed for Reasoning LLMs.   
\end{itemize}
All training scripts, prompts, and finetuning datasets of this work are publicly available~\citep{wetten25, wetten_OLMo, wetten_Mistral}.

\section{Related work}\label{related_work}

With the advent of LLMs, a new type of learning has emerged:  prompt-based (or in-context) learning~\citep{kamath_prompt-based_2024}. This type of learning occurs at inference time, using a structured prompt that includes a task description, optional examples, and a query. 
To enhance LLM reasoning, chain of thought (CoT) prompting was introduced, where the model is guided to generated intermediate steps before answering~\citep{wei2022chain}. 
In few-shot CoT prompts include a task or question, followed by a step-by-step reasoning example along with the final answer, and ending with a similar question or task. This approach showed better performance on complex reasoning tasks for large models.
%
%
\citet{kojima2022large} proposed a zero-shot CoT template for reasoning, 
they unlock the reasoning step by adding the {\em Let's think step by step} sentence at the end of each prompt. 

\citet{wang2023plan} introduce plan and solve prompting (PS), another zero-shot CoT method. It prompts the model to first plan a solution and then execute it, using the sentences {\em Let’s first understand the problem and devise a plan to solve the problem. Then, let’s carry out the plan and solve the problem step by step}. They extend these sentences with more detailed instructions to reduce errors in the reasoning step. 
%
Further approaches  on reasoning in LLMs can be found in  \citep{chu2023survey,dong2022survey,huang2022towards,plaat_reasoning_2024}.
Another approach to teaching a pretrained LLM to perform a new task is finetuning, where the parameters of the model are adjusted~\citep{jeong2024fine}. Finetuning adapts pretrained LLMs to specific task with smaller, domain-specific datasets. A common method is supervised finetuning (SFT), where labeled examples guide learning. Instruction-tuning, a variant of SFT, trains models on \textit{(Instruction, Output)} pairs, where  \textit{Instruction} is a human instruction and  \textit{Output} is the desired response by the LLM for that instruction~\citep{zhang_instruction_2024}.
Full finetuning updates all model parameters. This technique can be costly, as the pretrained model often contains billions of parameters, see early models such as GPT~\citep{radford_improving_2018}. Parameter-efficient finetuning (PEFT) offers a lighter alternative by modifying only a small subset of parameters~\citep{han_parameter-efficient_2024}. A notable PEFT method is LoRA~\citep{hu_lora_2021}, which inserts low-rank matrices to approximate weight updates, reducing memory and compute costs.


The rise of AI agents achieving dominance in gaming begins in the 1990s with Deep Blue~\citep{campbell2002deep}, which defeated world chess champion Garry kasparov using brute-force search and domain expertise.
%
Attention then shifted towards machine and reinforcement learning approaches \citep{silver2016mastering,plaat2020learning}. 
%
%
%
Currently,  agents and LLMs are converging \citep{plaat2025agentic}. 
In the field of AI agents playing games, ChessGPT~\citep{feng2024chessgpt} introduced a substantial game and language dataset for chess, upon which two models have been created. 
%
\citet{li_assessing_2024} evaluated LLMs in Minesweeper using different input formats, finding that GPT-4 outperformed GPT-3.5, although limitations remained.
%
\citet{Noever2021PuzzleSW} used GPT-2 to solve puzzles  such as mazes, Sudoku and the Rubik's Cube, by training on solved examples, demonstrating a text-based alternative to traditional search methods. 
These studies highlight the  potential of LLMs in solving puzzles and playing games.


\begin{figure}
    \begin{minipage}{.3\textwidth}
    \centering
    \begin{lstlisting}[basicstyle=\ttfamily\scriptsize]
            __________
            _B12...13_
            _........_
            _........_
            _.....F.._      
            _.b....f._
            _........_
            _........_
            _........_
            __________
    \end{lstlisting}
    \end{minipage}
    \begin{minipage}{.3\textwidth}
    \centering
    \includegraphics[width=.75\linewidth]{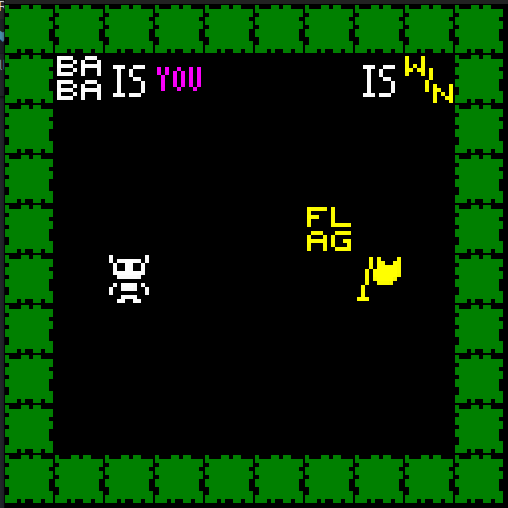}
    \end{minipage}%
    \begin{minipage}{.3\textwidth}
    \centering
    \includegraphics[width=.75\linewidth]{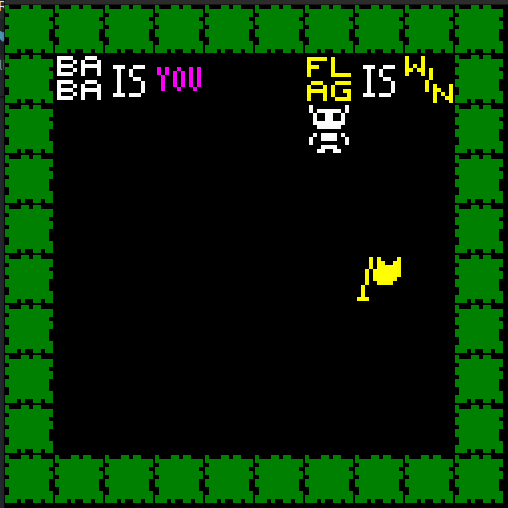}
    \end{minipage}%
    \captionof{figure}{ASCII representation of a level; Pictorial representation; By pushing (\textit{mention}), Baba has created the Rule (\textit{use}) FLAG IS WIN}
    \label{fig:baba1}
\end{figure}

\section{Method}\label{method}
\label{environment}
\begin{figure}
    \centering
    \begin{tabular}{c c c}
        \begin{subfigure}{0.4\textwidth}
            \centering
            \includegraphics[width=3cm]{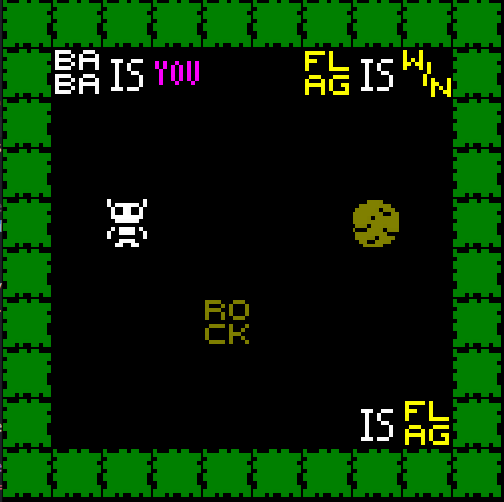}
            \caption{Level 4: A flag object is the win condition, but there is currently no flag object to reach}
        \end{subfigure} 
        & 
        \raisebox{3cm}{\(\Rightarrow\)} 
        & 
        \begin{subfigure}{0.4\textwidth}
            \centering
            \includegraphics[width=3cm]{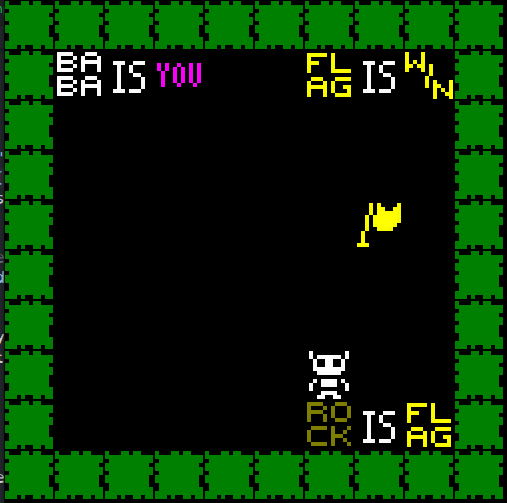}
            \caption{The rule ROCK IS FLAG is created and the rock object is transformed into a flag object}
        \end{subfigure} 
        \\
        \begin{subfigure}{0.4\textwidth}
            \centering
            \includegraphics[width=3cm]{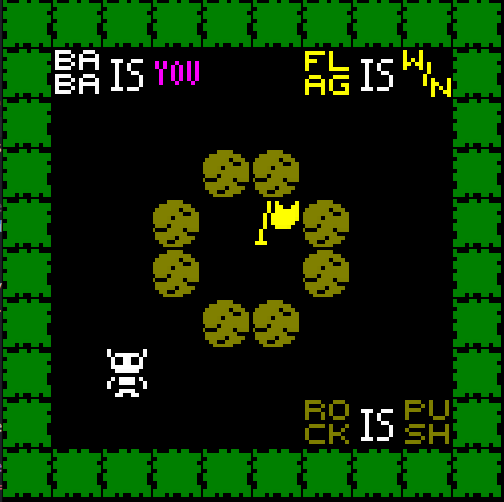}
            \caption{Level 6: The flag object, which serves as the win condition, is surrounded by pushable rock objects}
        \end{subfigure} 
        & 
        \raisebox{3cm}{\(\Rightarrow\)} 
        & 
        \begin{subfigure}{0.4\textwidth}
            \centering
            \includegraphics[width=3cm]{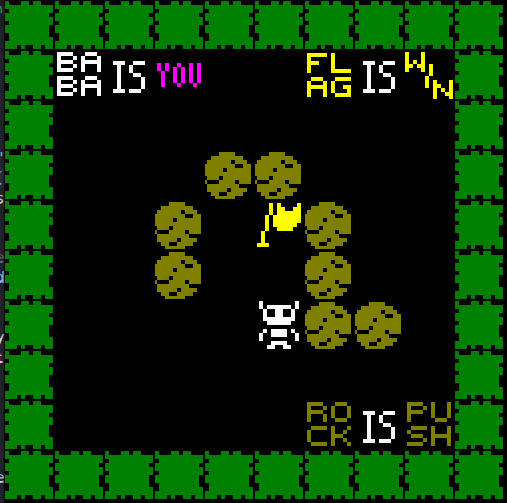}
            \caption{Two rock objects are pushed away by the baba object to clear the path to the flag object}
        \end{subfigure} 
        \\
        \begin{subfigure}{0.4\textwidth}
            \centering
            \includegraphics[width=3cm]{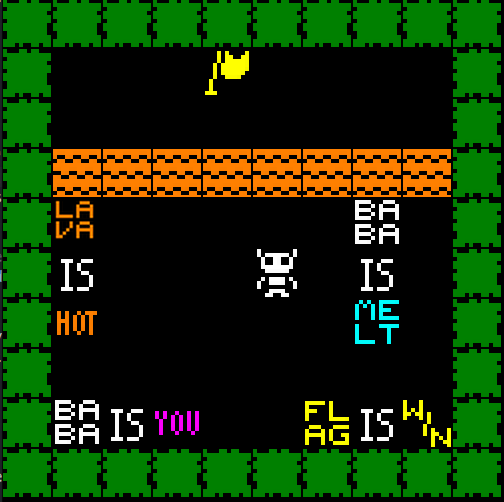}
            \caption{Level 12: The path towards the flag object (win condition) is blocked by lava because it is hot and baba object is set to melt}
        \end{subfigure} 
        & 
        \raisebox{3cm}{\(\Rightarrow\)} 
        & 
        \begin{subfigure}{0.4\textwidth}
            \centering
            \includegraphics[width=3cm]{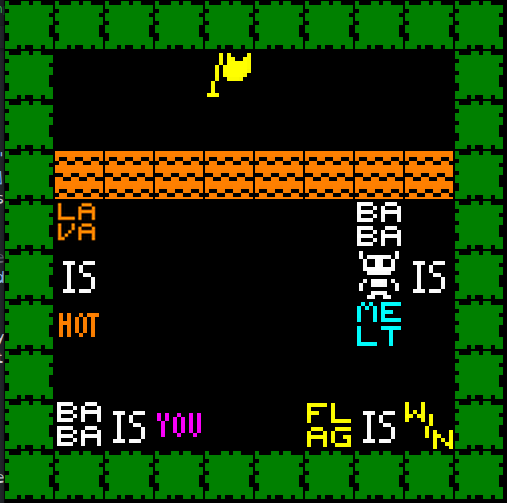}
            \caption{The rule BABA IS MELT is broken, this means baba will not melt when touching the lava, and the path toward the flag object is cleared}
        \end{subfigure} 
    \end{tabular}
    \caption{Examples of different game mechanics using the flag as a win condition. Each sequence shows how obstacles are manipulated or rules are changed to create a path to the flag.}\label{fig:examples}
\end{figure}

The game \textit{Baba is You}~\citep{teikari2019baba} is a 2D puzzle game with levels: a grid filled with objects and text blocks (see Figure~\ref{fig:baba1}). Text blocks can be used to create rules; these rules can be created from left to right or from top to bottom. The rule is active if there is at least one object, one verb, and one object or property aligned in a valid syntax  (Figure~\ref{fig:baba1}, right-most panel shows activation of the rule {\em FLAG IS WIN}). 
The primary components of the game are:
\begin{itemize}
    \item Objects: entities in the game, such as BABA, WALL, or ROCK.
    \item Verbs: These include IS and HAS, which create the backbone of syntactic rule construction.
    \item Properties: Attributes that determine how the object interacts within the environment, such as PUSH, STOP, WIN, or YOU.
\end{itemize}
%
%
Every solvable level should have a win condition and an object that can be controlled by the player. Rules dynamically define how objects behave in this game; an object does not really matter until there is a rule assigned to the object. 
Rules can be created, modified, or broken during gameplay by rearranging of text blocks. 
A level is considered complete when the object that is controlled by the player (IS YOU) touches the object designated as the win condition (IS WIN), or when the same object satisfies both rules. 
Figure~\ref{fig:examples} illustrates  gameplay scenarios.
The most common properties and rules are explained in Table~\ref{tab:rules} (Appendix).





In this work we used a simplified version of the game: the jam version of \textit{Baba is You}\footnote{\url{https://hempuli.itch.io/baba-is-you}} and a section of the game mechanics of the ``Game Module'' from Baba is Y'all~\citep{9231807}. 
Levels are encoded as strings of characters, where each character corresponds to an object or text block, see the ASCII grid representation in the left panel of  Figure~\ref{fig:baba1}.
%
In the simplified version of the game, the player is able to perform four distinct actions: 
\begin{itemize}
    \item Move: Navigate the controlled object towards other objects or text blocks;
    \item Create a Rule: Push text blocks into a valid rule by aligning them with a controlled object;
    \item Break a Rule: Break an active rule by pushing a text block away from its syntactic alignment;
     \item Push: Interact with text blocks, or objects if they are set to PUSH.
\end{itemize}

\subsection{Experimental Setup}
In this section we describe the methodologies used in our study, including data collection, experimental setup, and analysis techniques.
We evaluated six different LLMs. 
    \textit{GPT-4o}~\citep{hurst2024gpt} (OpenAI) is  a state-of-the-art reasoning model due to its strong performance on various tasks.  GPT-4o can reason at inference time~\citep{valmeekam_planning_2024}, using methods such as reinforcement learning to call a model multiple times with different prompts. 
    \textit{Gemini 1.5-Flash}~\citep{team_gemini_2024} (Google DeepMind) is designed for cost efficiency and fast inference.
    \textit{Mistral 7B instruct}~\citep{jiang_mistral_2023} (Mistral AI) is a 7B parameter model tuned for instruction-following tasks. \textit{Mixtral 8x7B}~\citep{jiang_mixtral_2024} uses a Mixture-of-Experts (MoE) design, activating only 12B of its 45B parameters per inference to reduce computational cost. 
    \textit{OLMo 7B and 13B instruct}~\citep{jiang_mistral_2023} (AI2) are open source models trained on the Dolma dataset, an open dataset that includes a mix of web content, academic publications, code, books, and encyclopedic materials. The OLMo models are designed with a focus on research accessibility, interpretability, and transparency.

To enable the model to understand and play the game \textit{Baba is You}, we constructed three different prompts. (Please refer to the Appendix.)
\begin{itemize}
    \item \textbf{Simple Prompt} 
    The prompt consists of the game mechanics, the definition of the characters to interpret the level, and the definition of each property.
    \item \textbf{Rule-extended Prompt} 
    Adds the rules that are active at the current level to the prompt.
    \item \textbf{Action-extended Prompt} 
    Further expands the prompt by including a description of the possible actions, partially adapted and extended from \citet{cloos2024baba}.
\end{itemize}
Each prompt ends with a question to solve the given grid level, followed by the ASCII grid level, and, at the end, a PS sentence (Plan-and-Solve \citep{wang_plan-and-solve_2023}) to activate CoT (Chain-of-Thought \citep{wei2022chain}). 
The prompts were constructed manually through iterative trial and error with GPT-4o. Outputs were reviewed for improvement, refined, and resubmitted until a satisfactory  version was achieved.
%
%
%
%
%
%

\begin{table}
\centering
\renewcommand{\arraystretch}{1.2}
\footnotesize
\begin{tabular}{m{2.7cm}|m{3.2cm}|m{7cm}}
\hline
\textbf{Category} & \textbf{Sub-Category} & \textbf{Description} \\ \hline

\textbf{Level Interpretation (w1)} & 
1 Hallucination & 
When information is completely out of context and not present in the levels \\ \cline{2-3}
& 2 Incorrect definitions & 
Incorrect classification of an object or text block or incorrect definition of a rule \\ \cline{2-3}
& 3 Incomplete Information & 
 Absence of defining objects, text blocks, or rules that are present in the level 
\\ \hline

\textbf{Formulate Problem Stmt (w2)} & 
1 Transfer of errors & 
Wrong problem statement derived from a previous incorrect statements \\ \cline{2-3}
& 2 Incomplete information & 
Missing elements in the problem statement \\ \cline{2-3}
& 3 Wrong assumptions & 
Assuming a condition or rule applies without explicit evidence \\ \cline{2-3}
& 4 Hallucination & When information is completely out of context and not present in the levels \\ \hline

\textbf{Formulating Solutions (w3)} & 
1 Transfer of errors & 
Wrong solution derived from a previous incorrect statements \\ \cline{2-3}
& 2 Hallucination & 
When information is completely out of context and not present in the levels \\ \cline{2-3}
& 3 Wrong reasoning & Drawing incorrect conclusions due to misinterpreting rules, neglecting constraints, or failing to account for rule interactions and ambiguities
 \\ \cline{2-3}
& 4 Incomplete solution & Missing steps to fully solve the level
 \\ \hline

\textbf{Formulating Actions (w4)} & 
1 Transfer of errors & 
Wrong actions derived from a previous incorrect statements \\ \cline{2-3}
& 2 Incomplete actions & 
Missing actions to complete the level \\ \cline{2-3}
& 3 Wrong format & 
The actions are in the wrong format (only prompt 3) \\ \cline{2-3}
& 4 Wrong actions & 
Actions proposed are not solving the level \\ \cline{2-3}
& 5 Hallucination & 
When actions are completely out of context \\ \hline

\end{tabular}
\caption{Error Categorization}\label{tab:error}
\end{table}

\subsubsection{Evaluation of Reasoning}\label{evaluation}
In order to investigate how well LLMs perform in reasoning and solving \textit{Baba is You} levels, a manual analysis was performed to examine the reasoning chains generated by LLMs. The reasoning chain can be divided into four distinct sections: the interpretation of the level, the formulation of the problem statement, the formulation of the solution for the problem, and the formulation of the actions that should be taken for the solution. The first two sections are part of the analysis of the level, while the latter two sections are part of the solution process so formulating an answer consists of four steps. Table~\ref{tab:error} summarizes errors encountered in these steps. The error categories are used in Figure~\ref{fig:prompt_based_error}. If a step is correct, it is marked with a {\em c} label together with the number of the step, otherwise errors are classified according to the subcategories. The correctness frequency is shown in Figure~\ref{fig:prompt_based_correct}.

\begin{figure}[t]
    \centering
    \begin{minipage}{0.18\linewidth}
        \centering
        \includegraphics[width=\linewidth]{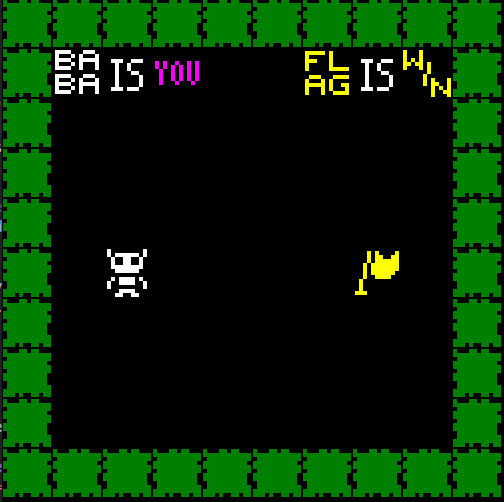}
        \caption*{(a) level 1} 
        \label{fig:level1}
    \end{minipage}\hspace{0.01\linewidth}
    \begin{minipage}{0.18\linewidth}
        \centering
        \includegraphics[width=\linewidth]{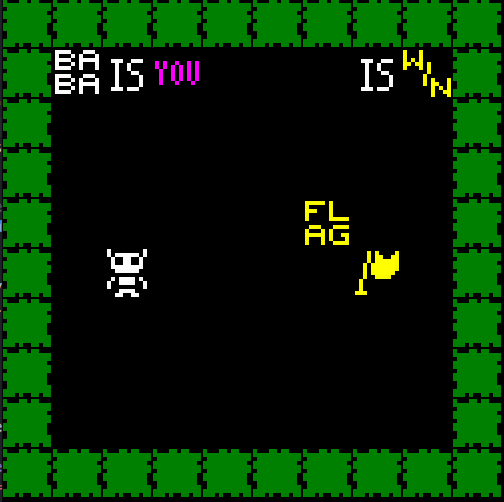}
        \caption*{(b) level 2} 
        \label{fig:level2}
    \end{minipage}\hspace{0.01\linewidth}
    \begin{minipage}{0.18\linewidth}
        \centering
        \includegraphics[width=\linewidth]{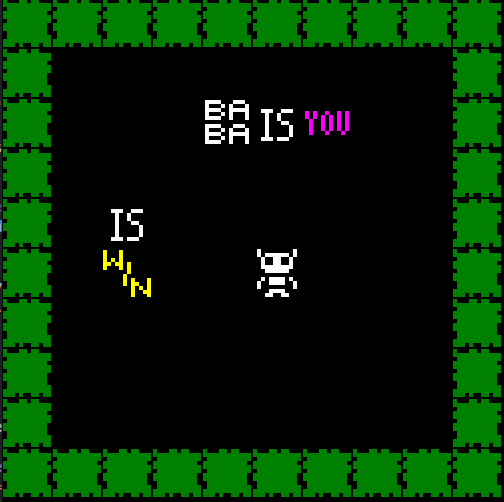}
        \caption*{(c) level 3} 
        \label{fig:level3}
    \end{minipage}\hspace{0.01\linewidth}
    \begin{minipage}{0.18\linewidth}
        \centering
        \includegraphics[width=\linewidth]{images/level4.png}
        \caption*{(d) level 4} 
        \label{fig:level4}
    \end{minipage}\hspace{0.01\linewidth}
    \begin{minipage}{0.18\linewidth}
        \centering
        \includegraphics[width=\linewidth]{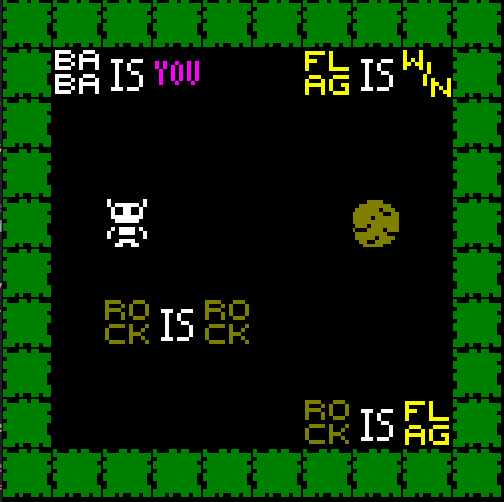}
        \caption*{(e) level 5} 
        \label{fig:level5}
    \end{minipage}

    \vspace{0.5cm} 

    \begin{minipage}{0.18\linewidth}
        \centering
        \includegraphics[width=\linewidth]{images/level6.png}
        \caption*{(f) level 6} 
        \label{fig:level6}
    \end{minipage}\hspace{0.01\linewidth}
    \begin{minipage}{0.18\linewidth}
        \centering
        \includegraphics[width=\linewidth]{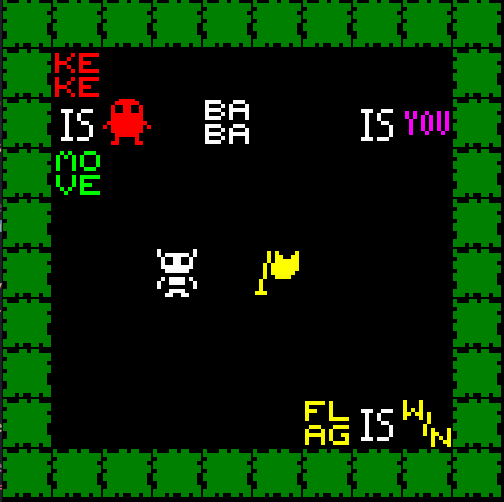}
        \caption*{(g) level 7} 
        \label{fig:level7}
    \end{minipage}\hspace{0.01\linewidth}
    \begin{minipage}{0.18\linewidth}
        \centering
        \includegraphics[width=\linewidth]{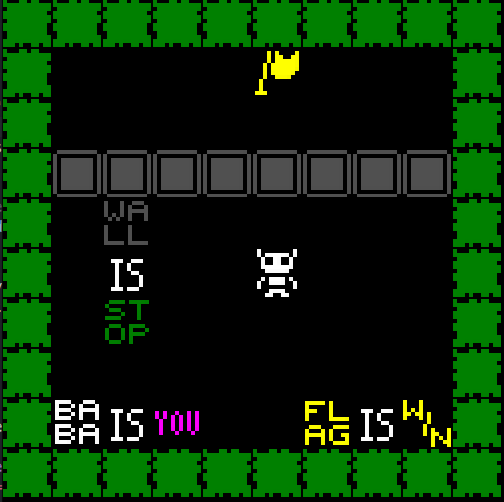}
        \caption*{(h) level 8} 
        \label{fig:level8}
    \end{minipage}\hspace{0.01\linewidth}
    \begin{minipage}{0.18\linewidth}
        \centering
        \includegraphics[width=\linewidth]{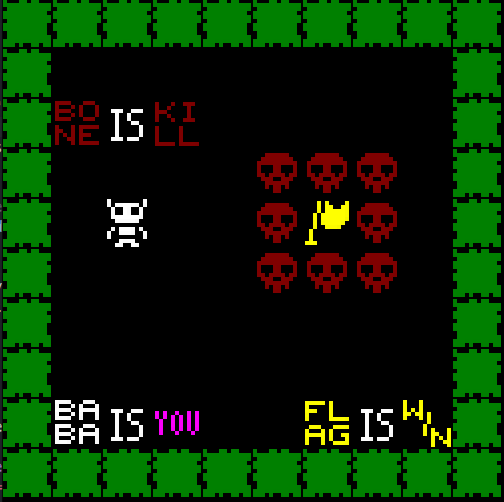}
        \caption*{(i) level 9} 
        \label{fig:level9}
    \end{minipage}\hspace{0.01\linewidth}
    \begin{minipage}{0.18\linewidth}
        \centering
        \includegraphics[width=\linewidth]{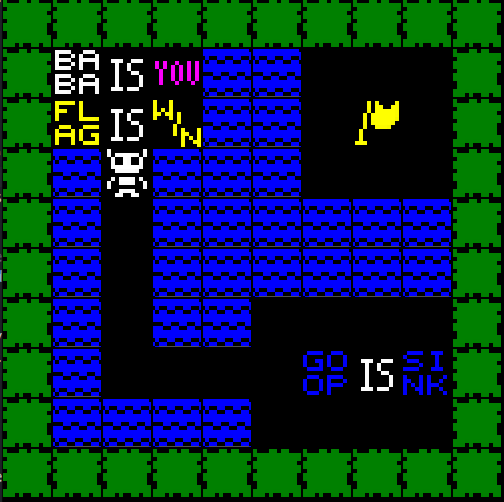}
        \caption*{(j) level 10} 
        \label{fig:level10}
    \end{minipage}

    \vspace{0.5cm} 

    \begin{minipage}{0.18\linewidth}
        \centering
        \includegraphics[width=\linewidth]{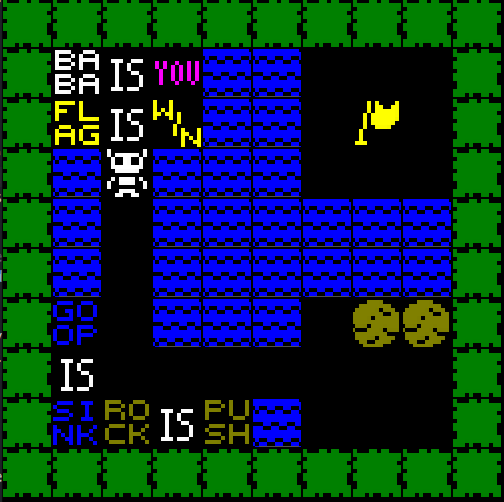}
        \caption*{(k) level 11} 
        \label{fig:level11}
    \end{minipage}\hspace{0.01\linewidth}
    \begin{minipage}{0.18\linewidth}
        \centering
        \includegraphics[width=\linewidth]{images/level12.png}
        \caption*{(l) level 12} 
        \label{fig:level12}
    \end{minipage}\hspace{0.01\linewidth}
    \begin{minipage}{0.18\linewidth}
        \centering
        \includegraphics[width=\linewidth]{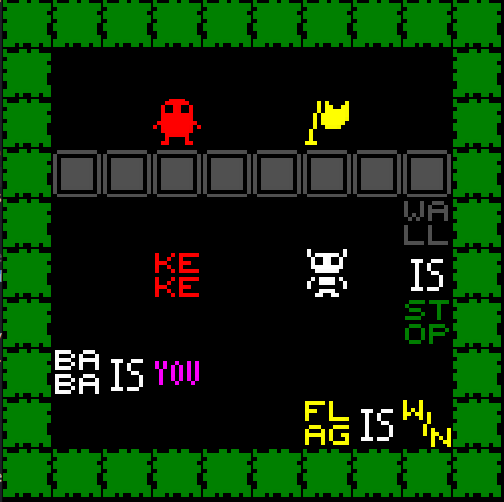}
        \caption*{(m) level 13} 
        \label{fig:level13}
    \end{minipage}\hspace{0.01\linewidth}
    \begin{minipage}{0.18\linewidth}
        \centering
        \includegraphics[width=\linewidth]{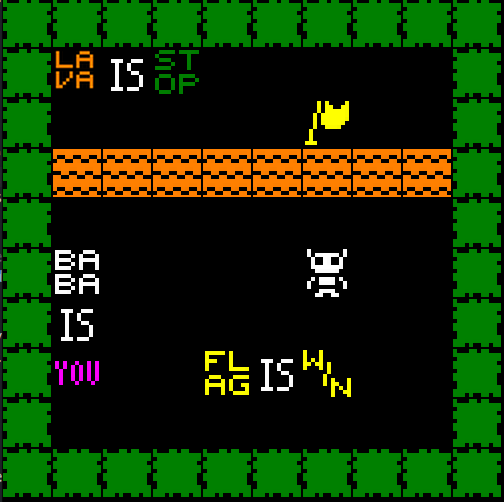}
        \caption*{(n) level 14} 
        \label{fig:level14}
    \end{minipage}

    \caption{Levels used for the evaluation of the LLM models in playing \textit{Baba is You}}
    \label{fig:levels}
\end{figure}

To evaluate the LLMs, each model and prompt format was tested in 14 different levels, each of which tests a different aspect of the game, see Figure~\ref{fig:levels}. Most of these levels are demo levels of the Keke AI competition~\citep{charity_keke_2022a}, except level 14. 
%
These levels require some logical thinking, but are relatively easy for humans, due to our natural ability to reason. However, LLMs encounter a challenge when confronted with the task of solving these levels. These models must not only interpret the rules and mechanics of the game from the text, but also apply them in the environment. Unlike humans, LLMs have no inherent understanding of the world. These models rely entirely on the information provided to decide how to interact with and manipulate the game state. The levels are designed to assess specific components of the game, including rule creation, transformation, immutability and logical reasoning, which are needed to determine the model's ability to play the game \textit{Baba is You}. 
%
%
%
%
%
%
%
%
Mistral and OLMo consistently produce identical outputs for repeated runs of the same prompt. In contrast, GPT-4o and Gemini Flash 1.5 exhibit variability. 
The accuracy was evaluated by running each prompt five times, a solution was correct if it appeared in at least three runs. 
%

For finetuning, we combined three different datasets. Each dataset contains a specific type of data.
The largest dataset~\citep{isaiahbjork} consists of various questions designed to improve the model's reasoning ability. These questions cover a range of logical and analytical challenges. The second dataset~\citep{wetten25} contains questions specifically related to the game mechanics of \textit{Baba is You}. It includes questions about the interactions between different game elements, the effects of specific rule changes, and the general logic of the game. The third dataset~\citep{wetten25} is the smallest and consists of different levels of \textit{Baba is You}. In this dataset, the input corresponds action-extended prompt of the level description, while the output represents the expected solution that the model should generate. Together, these three data sets form a combined data set used for finetuning, see Table~\ref{tab:dataset_samples} for the  sizes.
\begin{table}[t]
    \centering
    \begin{tabular}{|l|l|}
        \hline
        Dataset & Size \\
        \hline\hline
        CoT-logic-reasoning  & 10500 \\
        Questions  game mechanics & 289 \\
        Levels \& answers & 15 \\
        \hline

    \end{tabular}
    
    \caption{The size of the three datasets used for finetuning}
    \label{tab:dataset_samples}
\end{table}

The dataset containing questions about the game mechanics of \textit{Baba is You} was created through the following process. Initially, a set of questions was crafted, focusing on the rules and mechanics of the game. Then, GPT-4o was prompted to generate additional unique questions based on the ones we had already created. These generated questions and answers were reviewed and, when necessary, corrected. This iterative process allowed us to quickly build a solid dataset of questions related to the game's mechanics.
The other dataset, which consists of levels and their solutions, is entirely handwritten. As a result, this dataset is smaller, as more time was spent on creating detailed solutions for each level rather than on increasing the dataset size.
We trained Mistral 7B and OLMo 7B on the combined dataset using LoRA for parameter-efficient finetuning.

\section{Results}\label{results}
In this section, we present the findings of our study, analyzing the outcomes based on the predefined metrics. We start with the prompt-based learning results.
\subsection{Prompt-based learning}

\subsubsection{Simple prompt (1)}
The main challenge across models was to identify the active rules. GPT-4o performed better due to its improved recognition of objects and text blocks in the grid (Figure~\ref{fig:prompt_based_error}, prompt 2), though it still struggled with vertically placed rules. OLMo and Mistral had difficulty recognizing objects and text, which prevented them from formulating correct rules. As a result, most outputs for the simple prompt were incorrect (results are not shown).

\subsubsection{Rule-extended (2) and Action-extended  (3)}

\begin{figure}[t]
    \centering
    \includegraphics[width=0.99\linewidth]{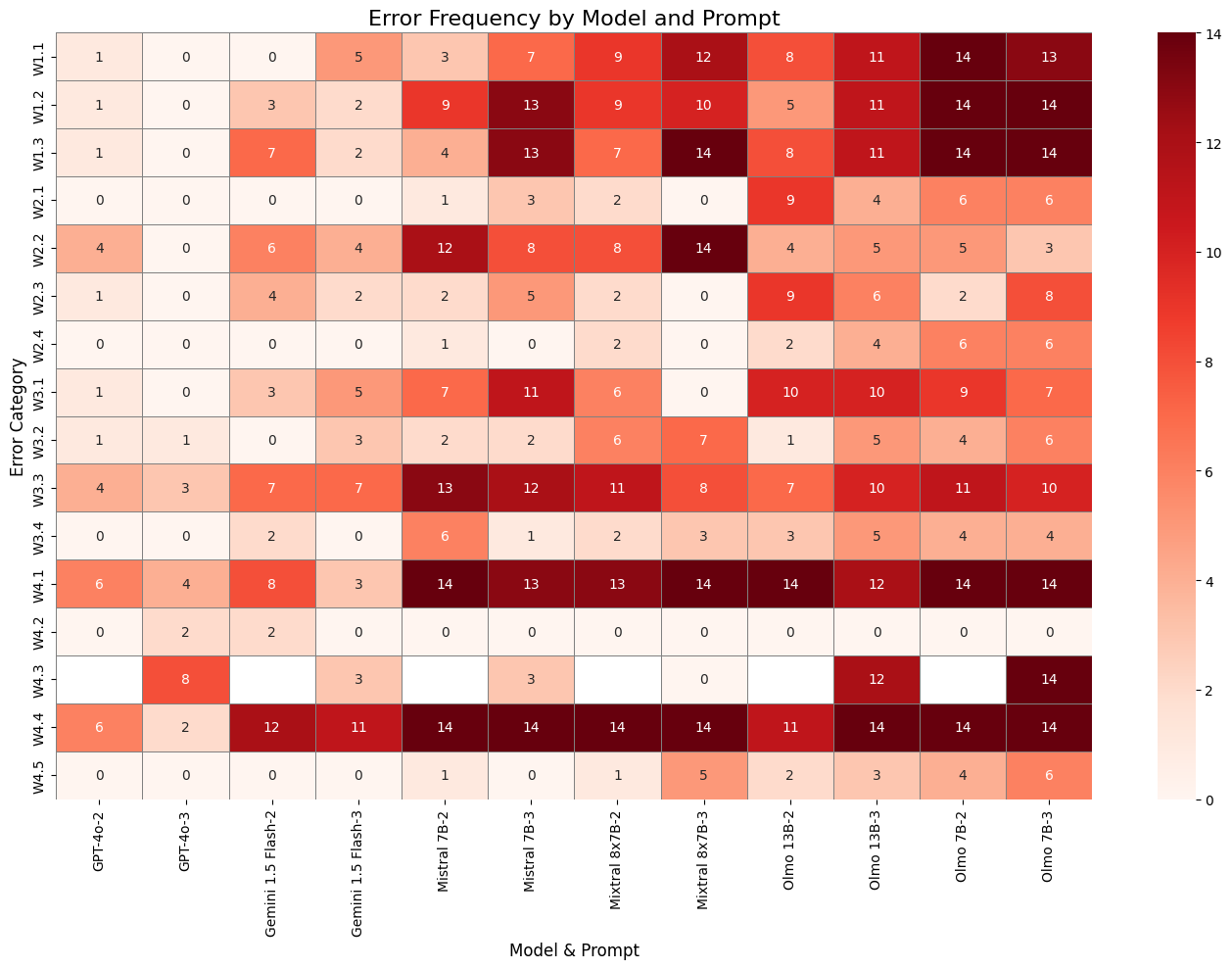}
    \caption{Frequency of error  step and subcategory (see Table~\ref{tab:error}) in the reasoning chains generated by the models. GPT-4o has the least errors. Furthermore,  OLMo 7B generates the most errors in the reasoning output for the levels, primarily when defining the text blocks and objects in the grid. Finally, all models encounter difficulties in formulating actions on the grid itself.}
    \label{fig:prompt_based_error}
\end{figure}

\begin{figure}[t]
\centering
    \includegraphics[width=0.6\linewidth]{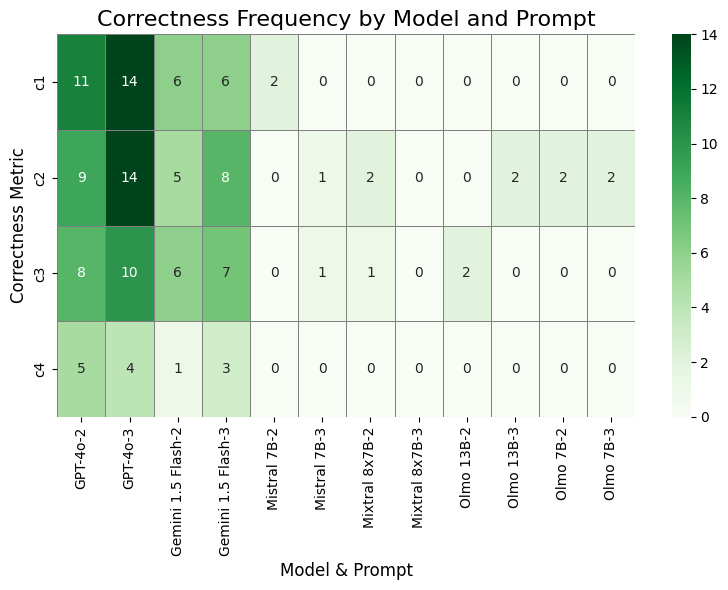}
    \caption{Correctness frequency per step in the reasoning chain generated by the LLM models (see Section~\ref{evaluation}). GPT-4o has generated the most correct steps in the reasoning chain for the levels. Furthermore,  GPT-4o and Gemini Flash 1.5 benefit from the action-extended prompt, while the smaller models encounter difficulties irrespective of prompt structuring. }
    \label{fig:prompt_based_correct}

\end{figure}




GPT-4o performs relatively well, demonstrating a strong ability to understand the grid and formulate correct problem statements (Figure~\ref{fig:prompt_based_error}). This LLM also provides reasonable level solutions (Figure~\ref{fig:prompt_based_correct}), making it potentially useful for assisting players. However, its action formulation is less reliable and often does not align with its own solutions.
%
%
A key weakness lies in distinguishing which rules are breakable. In levels 13 and 14, GPT-4o incorrectly suggests breaking unbreakable rules to reach the flag, highlighting a difficulty in understanding spatial constraints.
The action-extended prompt led to small improvements, with GPT-4o generating more accurate solutions and demonstrating better grasp of game mechanics. Still, it did not always strictly adhere to the action format.
Gemini Flash 1.5 performs slightly worse than GPT-4o in identifying objects and text in the grid, often leading to incomplete or occasionally missing problem statements. Although these issues were not a major obstacle for solution generation, the model struggled to consistently describe the obstacles. Also, with the third prompt, hallucinations increased (those in the first step did not transfer to other steps). There was also more hallucination in formulating the solution. 
Interestingly, at level 12 using prompt 2, the LLM proposed breaking 'BABA IS MELT' and forming 'LAVA IS MELT', causing lava to disappear, followed by creating 'BABA IS WIN', which results in a successful outcome. This solution is  intriguing because it is not the most straightforward approach, but a creative way to solve the puzzle.



\begin{figure}
    \centering
    \includegraphics[width=0.99\linewidth]{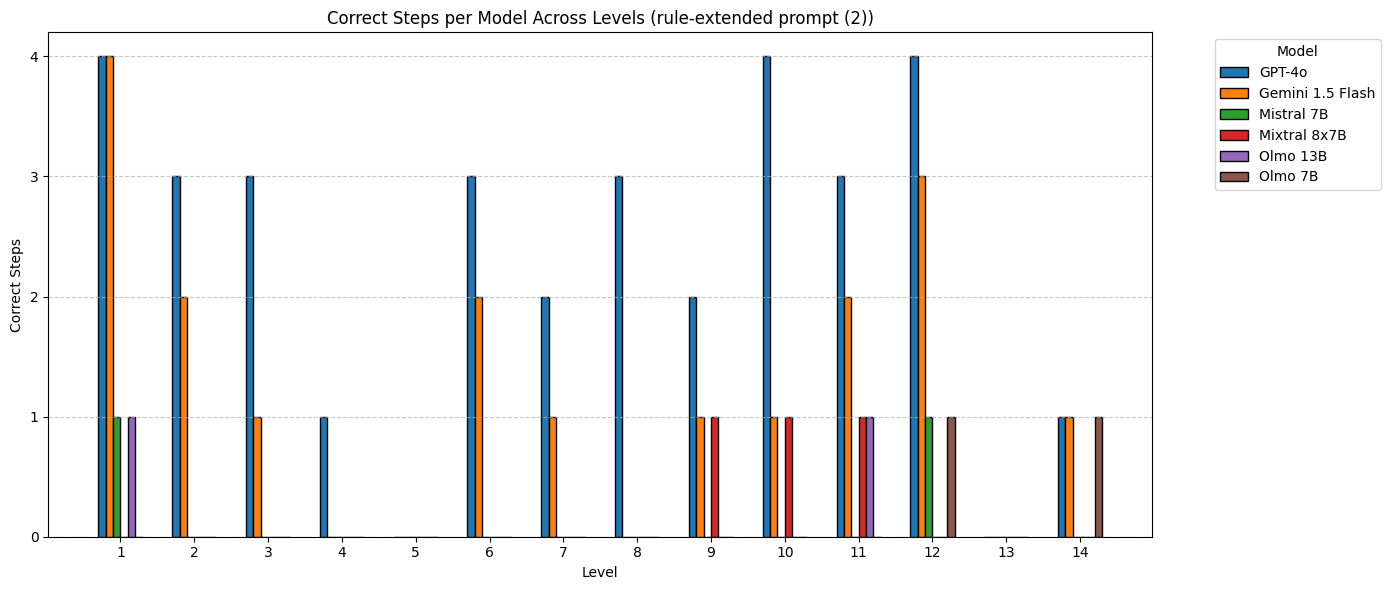}
    \caption{Correct steps per model across the 14 \textit{Baba is You} levels with the rule-extended prompt. The rule-extended prompt, which provides the active rules present in the level, improves performance across models but still highlights major differences in reasoning capabilities. GPT-4o outperforms other models, demonstrating stronger multi-step problem-solving skills. While Gemini 1.5 Flash show partial success, its performance remains inconsistent. The results suggest that simply providing active rules helps, but does not bridge, the gap in logical reasoning ability between smaller models and more advanced LLMs like GPT-4o. }
    \label{fig:levels_prompt2}
\end{figure}

\begin{figure}
    \centering
    \includegraphics[width=0.99\linewidth]{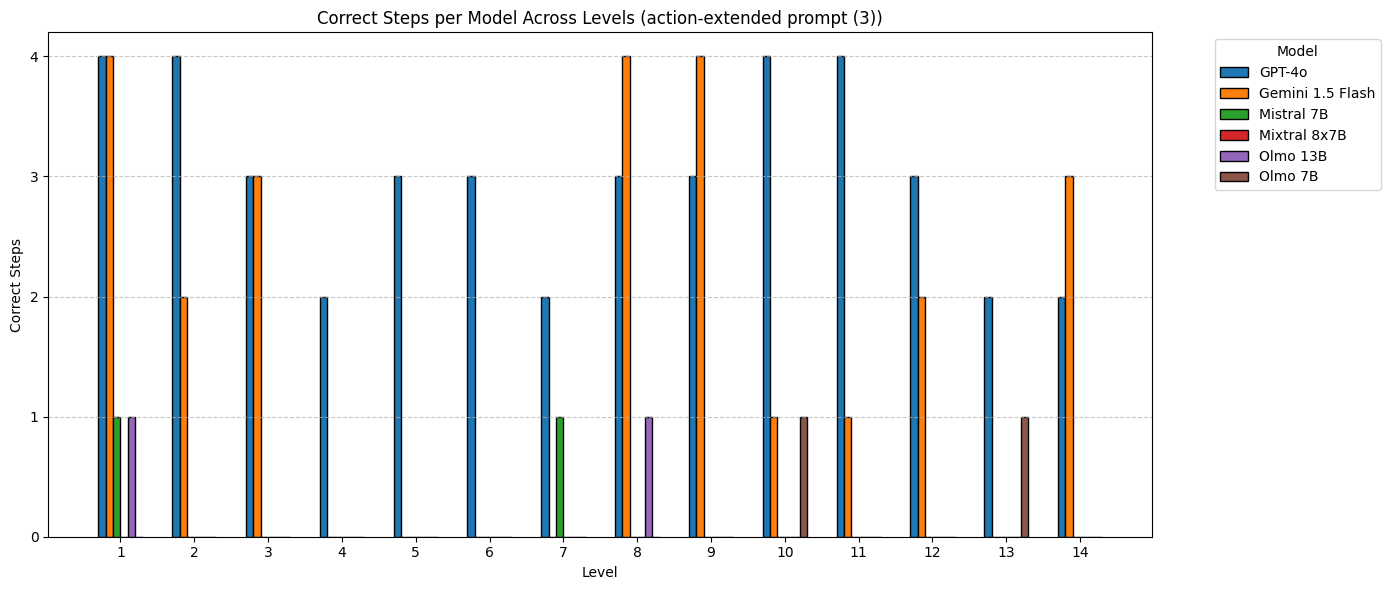}
    \caption{Correct steps per model across the 14 \textit{Baba is You} levels with the action-extended prompt. The action-extended prompt, which provides additional details about possible actions, leads to notable improvements for some models, particularly Gemini 1.5 Flash and GPT-4o. However, GPT-4o remains the strongest performer, consistently solving more steps across all levels. While some smaller models show slight improvements, their overall performance remains limited, suggesting that improving prompts alone is not sufficient to overcome their reasoning limitations. These results highlight the importance of both prompt design and underlying model capability in tackling complex rule-based reasoning tasks.}
    \label{fig:levels_prompt3}
\end{figure}

Like GPT-4o, Gemini Flash 1.5 struggles with generating accurate actions. Notably, it rarely suggested rule-breaking with the rule-extended prompt but did so more often with the action-extended prompt.
Overall, both models performed better with the action-extended prompt, showing fewer errors, and improved step-by-step reasoning (see Figures~\ref{fig:levels_prompt2} and~\ref{fig:levels_prompt3}). In Table~\ref{tab:llm_l_errors}  some examples of error snippets of the reasoning chain of GPT-4o and Gemini Flash 1.5 are shown (Appendix).
%
%
%
%
%
%
We also evaluated OLMo and Mistral models on 14 \textit{Baba is You} levels. Overall, they performed significantly worse than GPT-4o and Gemini Flash 1.5, particularly in object and text block identification in the grid (Figure~\ref{fig:prompt_based_error}). Table~\ref{tab:llm_errors} shows examples of error snippets of the reasoning chain of these models.
OLMo-7B and 13B struggled with hallucinations and frequent misidentification of grid elements. They often failed to distinguish between objects and rules, which led to incoherent problem statements and solutions. 
Both models frequently misinterpreted “WIN” as the target object and showed little understanding of the rule mechanics or grid constraints.

Mistral 7B and Mixtral 8x7B performed slightly better in object recognition but continued to produce flawed solutions. Mixtral often skipped the problem statement entirely and jumped straight to solutions, omitting context and causing logical gaps. A recurring issue was the models' misunderstanding of the rule-breaking mechanism. Rather than removing rules, they often suggested alternative rules, missing the mechanic’s intent. Additionally, both Mistral models struggled to track active rules, sometimes suggesting to create rules that already existed or were impossible to form given the available text blocks. One common error was assuming that movement required 'BABA IS MOVE,' indicating a lack of grasp of default game behavior.

\begin{figure}[t]
    \centering
    \includegraphics[width=0.8\linewidth]{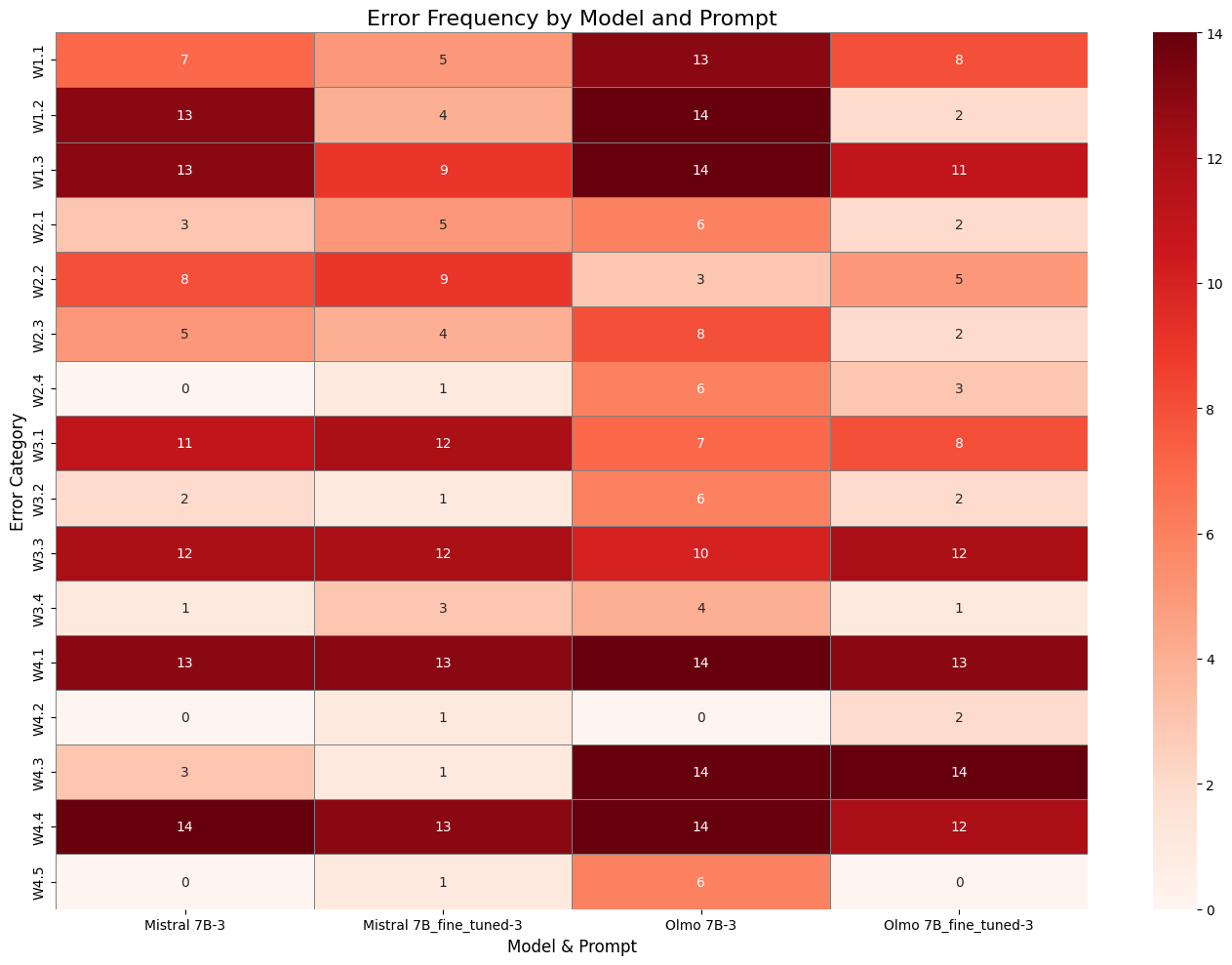}
    \caption{Frequency of errors per step and subcategory in the reasoning chains generated by the models when solving the \textit{Baba is You} levels. We see that both finetuned models have fewer errors in the analyzing part of the level in the reasoning chain after finetuning compared to the original model.}\label{fig:error_ft}
\end{figure}

\begin{figure}
    \centering
    \includegraphics[width=0.5\linewidth]{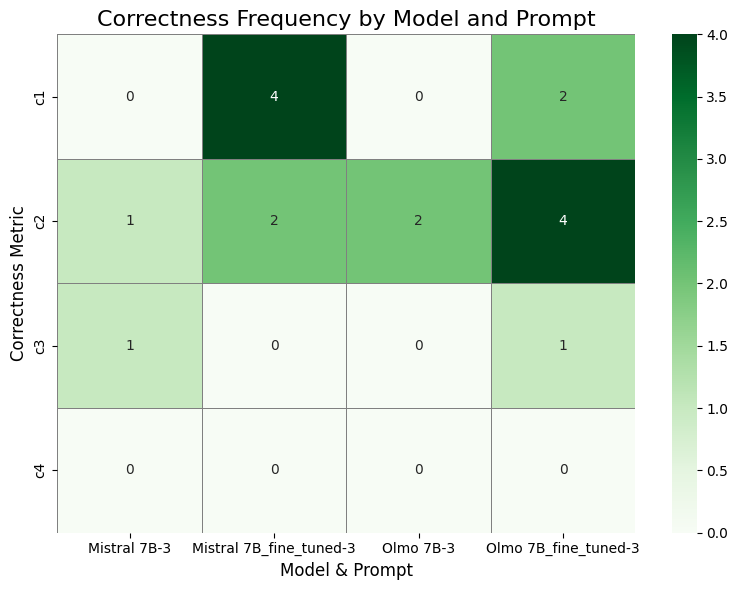}
    \caption{Correctness frequency per step in the reasoning chain generated by the LLM models. Both finetuned models have more correct steps after finetuning. For Mistral 7B there is an improvement in classification of objects and text blocks in the grid. For OLMo 7B there is an improvement in the problem statement formulation.}
    \label{fig:correct_ft}
\end{figure}


Unlike GPT-4o and Gemini 1.5 Flash, which showed improvements with structured prompts, OLMo and Mistral models did not consistently benefit from action-extended prompts. The solutions remained equally flawed.
These results highlight key limitations of smaller models: difficulty distinguishing game entities, tracking rule states, and reasoning through rule-breaking mechanics. While Mistral models showed slight improvement over OLMo, neither models demonstrated strong puzzle-solving ability.
%
%
%
In levels 4 and 5, most models misinterpreted the presence of the rule “FLAG IS WIN” as implying the flag’s existence, overlooking the need to create or transform the flag. In level 5, some models incorrectly assumed the flag was inside the rock due to “FLAG IS ROCK,” revealing confusion between rule-based transformation and object persistence.

\subsection{Finetuning}
\begin{figure}
    \centering
    \includegraphics[width=0.99\linewidth]{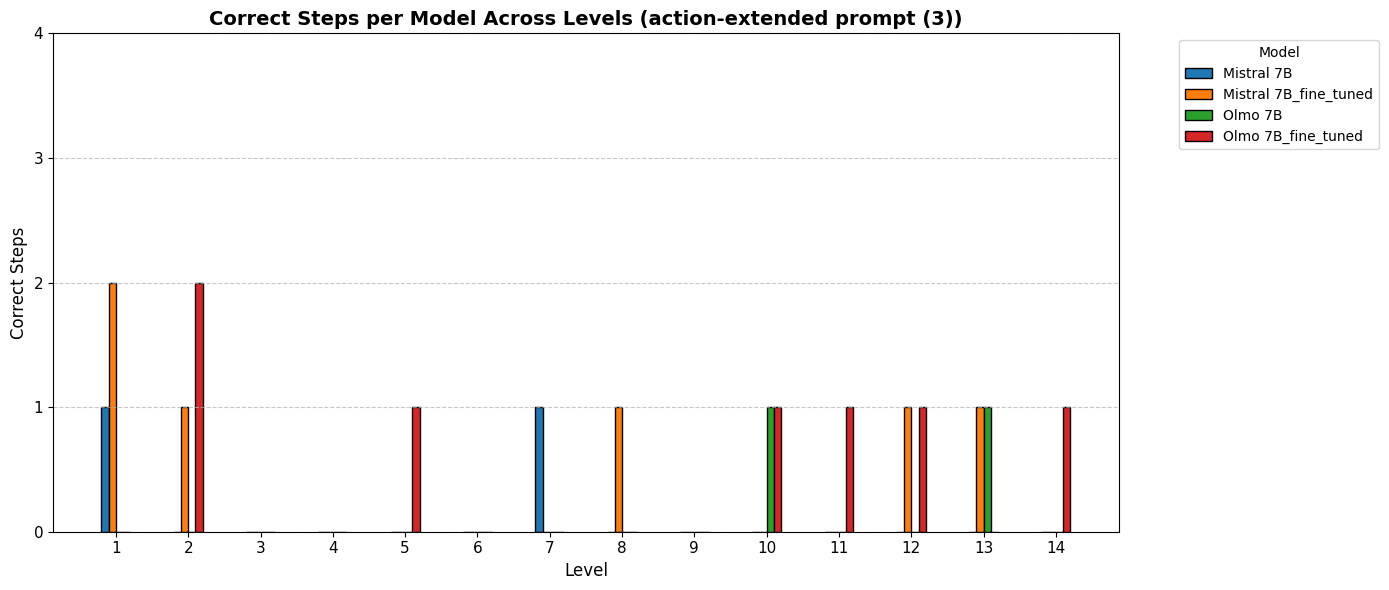}
    \caption{Correct steps per model across the 14 Baba is You levels with the action-extended prompt. The finetuned models have more correct steps across the levels but  still not enough to fully solve the levels.}
    \label{fig:levels_ft}
\end{figure}
Next, we  discuss the finetuning results on Mistral and OLMo.
Finetuning Mistral 7B improved  classification of the objects and text blocks in the grid (Figure~\ref{fig:correct_ft}). There were fewer misclassifications and incomplete information problems (Figure~\ref{fig:error_ft}). However,  this did not  improve problem statements or solving of  levels, which  was often incomplete with wrong assumptions  about grid-objects. 

After finetuning, OLMo 7B improved  the formulation of the problem statement (Figure~\ref{fig:correct_ft}) and achieved a reduction in classification errors for objects and text blocks (Figure~\ref{fig:error_ft}). However, the model still struggles with correctly distinguishing between them. 
The generated solutions suggest that the finetuned model still has difficulty grasping the game mechanics, rarely proposing actions such as breaking or creating rules. Additionally, it sometimes treats text blocks as the objects you control.
Finetuning the models with textual data from the game \textit{Baba is You} led to improvements in level analysis for both models. In the case of Mistral 7B, there was an improvement in classifying text blocks and objects, while for OLMo 7B, the problem statement formulation showed better results. However, for both models, there was no clear improvement in  solving the puzzle, as the generated solutions and actions still contained many errors.

\section{Discussion \& Conclusion}\label{conclusion}

In the puzzle game {\em Baba is You} the goal is to win by following rules and by creating new rules,  tasks that involve both language and reasoning abilities. LLMs must be able to move ({\em mention}) game pieces in such a way  that they align to form new rules ({\em use}). This study explores how various LLMs perform on 14 relatively simple game levels: how well they are able to solve levels  by understanding the consequences of rule manipulation and spatial understanding. We used prompt-based-learning first, finetuning second. Among the models evaluated with prompt-based-learning, GPT-4o and Gemini Flash 1.5 consistently outperform smaller models in identifying objects, interpreting game mechanics, and generating partially correct solution paths. However, even these more advanced models struggled to accurately interpret the grid as a two-dimensional space, often overlooking critical constraints such as rule-breakability.
%
%
Finetuning on structured textual data led to improvements, Mistral 7B showed better classification and OLMo 7B improved in formulating problem statements, but neither model demonstrated substantial gains in full solution generation. Mistral and OLMo continued to struggle with core aspects of the game such as distinguishing between text and object blocks and understanding how rule creation or breaking is physically performed in the game.

This work shows that while high-end models like GPT-4o and Gemini Flash 1.5 can reason through parts of \textit{Baba is You} levels, they still struggle to fully understand the game. Furthermore, without explicit prompts that include active rules and structured action formats, their performance drops significantly. A common limitation across all models is that they fail to interpret the grid as a two-dimensional space, leading to incorrect or overly simplistic solutions.
Even GPT-4o often fails to recognize which rules can be broken, and it is unclear whether LLMs truly grasp the mechanism of rule manipulation through moving text blocks. Smaller models such as Mistral and OLMo, even when finetuned, frequently misinterpret game elements and fail to demonstrate a solid understanding of the mechanics.

Complex reasoning tasks such as {\em Baba is You} pose three types of challenges to an LLM: challenges of (1) representation, (2) reasoning, and (3) reflection \citep{madaan2023self,schultz2024mastering,plaat_reasoning_2024}.

\textit{Representation} First, the LLM must be able to represent puzzle states correctly. In Chess, work on ChessGPT has shown that pretraining and finetuning can teach an LLM to recognize positions and solve problems correctly \citep{karvonen2024emergent,feng2024chessgpt}. In OthelloGPT, \citet{li2023emergent,nanda2023emergent} have used mechanistic interpretability to show how pretrained LLMs  represent boards internally. In  our study of {\em Baba is You}  LLMs were not pretrained on the game, and the LLMs have difficulty with the spatial interpretation of the board. Further finetuning and pretraining may be necessary for improvement.

\textit{Reasoning}
Second, in order to correctly manipulate the state representations, the LLM must be able to reason with the rules, for example, to follow chains of thought \citep{wei2022chain}.  \citet{schultz2024mastering,zhang2025complete} show that by pretraining and finetuning  on textual representations of Chess, LLMs can learn to reason  well enough to play correct games (although not yet at a high level of play). In {\em Baba is You}, we also saw that finetuning was able to enhance reasoning.

\textit{Reflection}
Third, in {\em Baba is You} the LLM must be able to reflect on its own reasoning to understand the effect ({\em use}) of the rules that it composes ({\em mention}). Reasoning LLMs typically apply reinforcement learning to reflect on their own actions, using an external algorithm to control the self-reflection process \citep{madaan2023self,shinn2023reflexion,yao2023tree}. In {\em Baba is You}, the LLMs achieve weak use-mention-type reasoning about dynamic rules, with prompts based on Plan-and-Solve \citep{wang_plan-and-solve_2023}. Achieving accurate use-mention reflection in {\em Baba is You}  may  require such explicit  external algorithms or methods such as analogical prompting \citep{yasunaga2023large}.



\subsubsection{Limitations \& Further work}\label{limitations}

Chain of thought (CoT) prompting~\citep{wei2022chain} has spawned active research in  methods for reasoning.
This research used plan and solve \citep{wang_plan-and-solve_2023}, a zero-shot CoT prompting method. Further research may explore other prompting methods, for example using explicit step-by-step prompting~\citep{shinn2023reflexion,press_measuring_2023,madaan2023self}. 

The evaluation was conducted on  14 relatively simple levels, which may not reflect model performance on more challenging puzzles. More levels, with varying difficulty, could provide deeper insights.
Furthermore, the finetuning dataset that we used was small, with only 15 examples for level solutions and 298 game mechanics questions. This restricts the model's exposure to the game's complexity. Expanding the dataset with more varied levels, solution paths, and mechanism-related questions may improve generalization and reasoning performance.
In general, larger models tend to perform better at test time inference~\citep{muennighoff2023scaling}, as the performance of GPT-4o in our work also indicated. Therefore, especially for reasoning and reflection, further research with larger models is warranted.
%
 Additionally, more extensive finetuning with adjusted hyperparameters (e.g., learning rate, batch size, or epochs) might yield better results.
%
Finally, error analysis in this work was performed manually, introducing potential subjectivity. Future work could implement automated evaluation tools to ensure more consistent and scalable assessment.



{\small
\bibliographystyle{plainnat}
\bibliography{bibliography, references}
}
\newpage

\appendix

\section{Prompts and Error Snippets}

\begin{table}[ht]
    \centering
    \begin{tabulary}{\linewidth}{|C|L|}
    \hline
        $<$object 1$>$ IS $<$object 2$>$ & Transforms all instances of object 1 into object 2\\ \hline
         $<$object 1$>$ IS $<$object 1$>$ & Set an object to be itself and therefore become immutable for the transformation rule.\\\hline
         $<$object$>$ IS WIN & Makes the object the win condition; anything controlled by you that touches or is the object wins. \\\hline
         $<$object$>$ IS YOU & Objects that are set to be YOU can be controlled by you. All objects will move simultaneously. \\\hline
         $<$object$>$ IS PUSH & Objects set to PUSH can be moved by either an object controlled by you or by other movable objects, shifting them one tile forward and occupying the new tile if possible. \\\hline
         $<$object$>$ IS MOVE &  Objects move one tile per turn in their facing direction, reversing direction if they hit an object, and if an object is both YOU and MOVE, it moves two tiles per turn.\\\hline
         $<$object$>$ IS STOP & Objects becomes impassible, important is that the nature of the PUSH rule overrides the blocking mechanism. \\\hline
         $<$object$>$ IS SINK & Objects that are set to SINK will destroy themselves as well as any other object that comes in contact with them. \\\hline
         $<$object$>$ IS HOT, $<$object$>$ IS MELT & Objects set to HOT will destroy anything set to MELT that touches them. If an object is both HOT and MELT, it will destroy itself. \\\hline
         $<$object$>$ IS KILL & Objects set to KILL destroy all objects controlled by you when they touch them, while remaining intact themselves. Note that the nature of the PUSH rule overrides this mechanic. \\\hline
    \end{tabulary}
    \caption{Explanation of common rules in the game BABA IS YOU}
    \label{tab:rules}
\end{table}

\begin{table}[hp]
    \centering
    \scriptsize
    \begin{tabular}{|l|p{10cm}|}
        \toprule
        \textbf{LLM Model} & \textbf{Error Snippets} \\
        \midrule
        \multirow{3}{*}{GPT-4o} & \textbf{Level 13: } \texttt{Goal:}\newline
        \texttt{To win the level, BABA (controlled by the player) must reach the FLAG (f). However, the WALL blocks the direct path.} \newline
        \texttt{Strategy:} \newline
        \texttt{Modify the rules to bypass or remove the obstacle created by the WALL:
        \textcolor{red}{Remove or break the "WALL IS STOP" rule.}}\newline
        \texttt{\textcolor{red}{Create an accessible path for BABA to reach the FLAG.}}\\ \cmidrule{2-2}
            &\textbf{Level 4: } \texttt{Identify Barriers:} \newline
        \textcolor{red}{\texttt{BABA (controlled by the player) must reach the FLAG, but there is no direct path. ROCK (r) is blocking BABA’s movement, and it has no properties assigned.}\newline
        \texttt{Use Rules to Clear the Path:}\newline
        \texttt{Assign ROCK IS PUSH by moving the text blocks R, 1, and 5 (if present) to form the rule.}\newline
        \texttt{This will allow BABA to push the ROCK (r) out of the way.}\newline
        \texttt{Reach the FLAG:}\newline
        \texttt{With the ROCK out of the way, BABA can move directly to the FLAG.}} \\

        \midrule
        \multirow{3}{*}{Gemini 1.5 Flash } &\textbf{Level 4:}\texttt{**Objects:** We have Baba (b), a Rock (r), and \textcolor{red}{Flags (f).}}\newline
        \texttt{**Text Blocks:** BABA (B), IS (1), YOU (2), FLAG (F), and WIN (3).}\newline
        \texttt{**Current Rules:** BABA IS YOU (you control Baba) and FLAG IS WIN (reaching a flag wins the level).}\newline
        \texttt{**Plan**}\newline
        \textcolor{red}{\texttt{Currently, the "FLAG IS WIN" rule is broken because the IS (1) and WIN (3) are not next to the FLAG (F). We need to move the IS and WIN next to the FLAG to make the rule active again.}}  \\ \cmidrule{2-2}
                        &\textbf{Level 5:} \texttt{Plan}\newline
                \texttt{The goal is to reach the Flag. \textcolor{red}{However, the initial rules make all Rocks into Flags, which can be confusing and potentially block the path. We need to change the rules to make the Flag a win condition again and allow Baba to move freely.}}\\ \cmidrule{2-2}
                    &\textbf{Level 13:}\texttt{**Solution**}\newline
                \texttt{1. **Breake\_Rule[GOOP IS SINK]:** Remove the rule "GOOP IS SINK." This will prevent the Goop from destroying itself.}\newline
                \textcolor{red}{\texttt{2. **Make\_Rule[ROCK IS HOT]:** Create the rule "ROCK IS HOT." This will make the Rocks hot.}\newline
                \texttt{3. **Make\_Rule[GOOP IS MELT]:** Create the rule "GOOP IS MELT." This will make the Goop melt when it touches the hot Rocks.}} \\
        \bottomrule
    \end{tabular}
    \caption{Examples of error snippets in the reasoning chains of GPT-4o and Gemini 1.5 Flash LLM models, where the red-colored text indicates incorrect reasoning. These errors highlight challenges such as hallucinations, incorrect assumptions, and overlooking key constraints.}
    \label{tab:llm_l_errors}
\end{table}

\begin{table}[hp]
    \centering
    \scriptsize
    \begin{tabular}{|l|p{10cm}|}
        \toprule
        \textbf{LLM Model} & \textbf{Error Snippets} \\
        \midrule
        \multirow{3}{*}{Mistral 8x7B} &\textbf{Level 1:} \texttt{Step 3 Devise a plan} \newline
                    \texttt{- We need to move the BABA object to the FLAG object.} \newline
                    \textcolor{red}{\texttt{- To do this, we need to change the rule BABA IS YOU to BABA IS MOVE.}\newline
                    \texttt{- Then, we need to push the ROCK object to the WALL object to create a path for the BABA object to reach the FLAG object.}} \\ \cmidrule{2-2}
                &\textbf{Level 3:} \texttt{Step 1: Understand the problem and extract the relevant objects, text blocks, and rules.}\newline
                \texttt{Objects: BABA, \textcolor{red}{FLOOR, GRASS, WALL, SKULL}}\newline
                \texttt{Text Blocks: 1, 2, 3}\newline
                \texttt{Rules: BABA IS YOU}\newline
                \textcolor{red}{\texttt{Step 2: Devise a plan to solve the problem.}\newline
                \texttt{1. Make BABA IS YOU to BABA IS MOVE.}\newline
                \texttt{2. Move BABA to the position of the SKULL.}\newline
                \texttt{3. Make SKULL IS YOU.}}\\ \cmidrule{2-2}
            &\textbf{Level 4: } \texttt{Step 2: Extract the relevant objects, text blocks and rules}\newline
        \texttt{- Objects: BABA, \textcolor{red}{SKULL, FLAG, FLOOR, GRASS,} ROCK, \textcolor{red}{WALL, KEKE, GOOP, LOVE,} and empty space.}\newline
        \texttt{- Text blocks: BABA, IS, YOU, FLAG, IS, WIN.}\newline
        \texttt{- Rules: BABA IS YOU, FLAG IS WIN.}\newline
        \texttt{Step 3: Devise a plan}\newline
        \textcolor{red}{\texttt{- Make BABA the object you control.}\newline
        \texttt{- Make FLAG the object you need to reach to win the level.}}\\

        \midrule
        \multirow{3}{*}{Mistral 7B } &\textbf{Level 1:} \texttt{The goal is to move the BABA to the FLAG to win the level.}\newline
                            \texttt{Plan:}\newline
                            \textcolor{red}{\texttt{1. Move the BABA to the flag by pushing the blocks in the way.}\newline
                            \texttt{2. Make the BABA move towards the flag.}\newline
                            \texttt{3. Make the BABA push the flag towards itself.}}\\ \cmidrule{2-2}
                        &\textbf{Level 3:} \texttt{Plan:}\newline
                    \textcolor{red}{\texttt{1. Move BABA to the 1 text block.}\newline
                    \texttt{2. Change BABA into the flag (f) using the 1 text block.}\newline
        \texttt{3. Move the flag to the exit (the empty space next to the 3 text block).}}\\ \cmidrule{2-2}
                    &\textbf{Level 9:}\newline
                    \textcolor{red}{\texttt{The goal is to move the flag to the exit (position B12) without getting killed by the skull.}}\\
        \midrule
        \multirow{3}{*}{OLMo 13B } 
                                &\textbf{Level 3:} \texttt{**Objective:**}\newline
                                \textcolor{red}{\texttt{Reach the WIN block to win the level.}}\\\cmidrule{2-2}
                                &\textbf{Level 4:}\texttt{**Plan:**}\newline
                       \textcolor{red}{\texttt{1. **Make Baba movable:** Ensure Baba has the "MOVE" property.}}\\ \cmidrule{2-2}
                        &\textbf{Level 9:} \texttt{1. **Neutralize SKULL:**}\newline
                        \textcolor{red}{\texttt{- We cannot directly push SKULL because it will kill BABA.}\newline
                       \texttt{ - We need to use text blocks to change SKULL's properties.}\newline
                    \texttt{- **Make Rule:** SKULL IS STOP. This will make SKULL impassable and safe for BABA.}}\\

        \midrule
        \multirow{3}{*}{OLMo 7B } &\textbf{Level 1:}\texttt{ \#\#\# Plan}\newline
                            \texttt{1. **Identify the Goal:** Move BABA to the FLAG \textcolor{red}{(F13).}}\newline
                            \textcolor{red}{\texttt{2. **Current Obstacles:** BABA starts at B12, and there are walls (W) and empty spaces (.) blocking the path.}}\\ \cmidrule{2-2}
                            &\textbf{Level 2:} \texttt{ 2. **Initial Setup:**}\newline
                            \textcolor{red}{\texttt{- BABA is currently next to a WALL (W) and a KEKE (K).}\newline
                            \texttt{- There's a FLAG (F) adjacent to BABA.}}\\ \cmidrule{2-2}
                            &\textbf{Level 9:}\texttt{ \#\#\# Plan:}\newline
            \textcolor{red}{\texttt{1. **Move BABA away from SKULL:** To avoid immediate death, move BABA away from SKULL.}\newline
            \texttt{2. **Create a path to FLAG:** Use the environment to create a path to FLAG.}\newline
            \texttt{3. **Utilize GOOP:** GOOP can be used to block SKULL or create a bridge}}\\

        \bottomrule
    \end{tabular}
    \caption{Examples of error snippets in the reasoning chains of OLMo and Mistral LLM models, where the red-colored text indicates incorrect reasoning. These errors highlight challenges such as hallucinations, incorrect assumptions, and wrong reasoning steps.}
    \label{tab:llm_errors}
\end{table}

\begin{figure}[H]
\begin{tcolorbox}[ exactbox, title={\textbf{Model Input}}]

You are helping to solve a gridworld game. In \textit{Baba is You}, the player can change the game rules by moving text blocks around. The grid contains object text and property text.

Text Blocks:\\
Object Text: Words representing game objects.\\
Property Text: Words that describe actions or properties.

Active Rules:\\
A rule is formed when the text blocks are aligned in a valid way, either horizontally or vertically. Valid rule formats include:\\  
\texttt{<object\_text> IS <property\_text>:} Grants a property to an object.\\
\texttt{<object1\_text> IS <object2\_text>:} Changes one object into another.\\
\texttt{<object1\_text> IS <object1\_text>:} Makes an object immutable.  

The goal is to use these rules to solve the level by moving the text blocks and controlling the objects.

\medskip
The level is displayed in a 2D grid, where each character has the following meaning:\\
Text blocks in the game which always can be pushed:\\
\texttt{<object\_text>:}\\
B = BABA, S = SKULL, F = FLAG, O = FLOOR, A = GRASS, L = LAVA, R = ROCK, W = WALL, K = KEKE, G = GOOP, V = LOVE \\

\texttt{<property\_text>:}\\
1 = IS, 2 = YOU, 3 = WIN, 4 = KILL, 5 = PUSH, 6 = STOP, 7 = MOVE, 8 = HOT, 9 = MELT, 0 = SINK  

\medskip
Objects in the game:\\
\texttt{<object>:}\\
b = object baba, s = object skull, f = object flag, o = object floor, a = object grass, l = object lava, r = object rock, w = object wall, k = object keke, g = object goop, v = object love, \_ = border, . = empty space  

\medskip
\texttt{<object\_text> IS YOU:} Makes the object you control in the game. You can move it and push blocks.\\
\texttt{<object\_text> IS WIN:} The object you need to reach or to be to win the level.\\
\texttt{<object\_text> IS STOP:} Makes the object impassable (you can't move through it).\\
\texttt{<object\_text> IS MELT:} Makes the object melt when touched by something marked as HOT.\\
\texttt{<object\_text> IS HOT:} Makes the object destroy any object marked as MELT when they touch it. If the same object is both HOT and MELT it self-destructs.\\
\texttt{<object\_text> IS MOVE:} Makes the object move one step in its direction every turn.\\
\texttt{<object\_text> IS KILL:} Makes the object destroy anything you control when touched, but it stays intact.\\
\texttt{<object\_text> IS PUSH:} Lets you push the object or have it pushed by other moving objects.\\
\texttt{<object\_text> IS SINK:} Makes the object destroy itself and anything it touches when it is touched.

\medskip
\textbf{Question:} Give a solution to the following grid level:
\begin{lstlisting}[basicstyle=\ttfamily\tiny]
__________\n_B12..F13_\n_........_\n_........_\n_........_\n_.b....f._\n
_........_\n_........_\n_........_\n__________\n
\end{lstlisting}
Let's first understand the problem, extract the relevant objects, text blocks and rules (explain the rules) in the level and make a plan to solve the problem. Then let's carry out the plan by giving the intermediate actions (using common sense). Solve the problem step by step and show the solution.

\end{tcolorbox}
\caption{\textbf{Simple prompt}: consisting only of a short game description and definitions of the characters and rules. Followed by a question to solve a level with at the end a sentence to activate zero-shot CoT.}\label{fig:first_prompt}
\end{figure}

The Rule-extended prompt and the Action-extended prompt can be found on GitHub~\citep{wetten25, wetten_OLMo, wetten_Mistral}.

\end{document}